\documentclass[a4paper]{article}

\usepackage{INTERSPEECH2021, caption, amsmath, graphicx, url, algorithm, algorithmic}
\title{Noisy student-teacher training for robust keyword spotting}
\name{Hyun-Jin Park, Pai Zhu, Niranjan Subrahmanya, Ignacio Lopez Moreno}

\address{
  Google Inc.
  }
\email{\{hjpark,paizhu,sniranjan,elnota\}@google.com}

\begin{document}

\maketitle
\begin{abstract}
We propose self-training with noisy student-teacher approach for streaming keyword spotting, that can utilize large-scale unlabeled data and aggressive data augmentation.
The proposed method applies aggressive data augmentation (spectral augmentation) on the input of both student and teacher and utilize unlabeled data at scale, which significantly boosts the accuracy of student against challenging conditions. Such aggressive augmentation usually degrades model performance when used with supervised training with hard-labeled data. Experiments show that aggressive spec augmentation on baseline supervised training method degrades accuracy, while the proposed self-training with noisy student-teacher training improves accuracy of some difficult-conditioned test sets by as much as 60\%.
\end{abstract}


\section{Introduction}
\label{sec:intro}

Supervised learning has been the major approach in keyword spotting area \cite{MaxPool20, Alvarez2019, He2018StreamingES, Cascade17, Custom17, Cnn15, Alexa18, AlexaDelayed18, Alexa17, Alexa16, HeySiri17}.
Although it has been successful, Supervised learning requires high quality labeled data at scale, which often requires expensive human efforts or costs to obtain. 
Motivated from such difficulty, Semi-supervised learning \cite{CPC2018, NeurIPS2019, wav2vec2019, SemiSupKWS2019, wav2vec2020} and Self-training \cite{SelfTrainImage20, SelfSupervised19, FewShot19, SelfTrainDomainAdapt19, SelfTrainSeqGen19} approaches were introduced recently. Those approaches utilize large unlabeled data in addition to smaller amounts of labeled data and achieve performance comparable to supervised models trained with large amounts of labeled data.

Semi-supervised learning approaches utilize unlabeled data to learn hidden layer activations that best predict neighboring data (temporally or spatially close sensory features), which is then used as the input feature to a classification network trained with small amount of labeled data (supervised training) \cite{wav2vec2019, NeurIPS2019, SemiSupKWS2019, wav2vec2020}. It is shown that semi-supervised learning with small amount of labeled data can achieve performance comparable to supervised learning with larger amount of data. 
On the other hand, Self-training approaches utilize unlabeled data by using a teacher network to generate soft-labels (pseudo-labels) which is then used to train student network \cite{SelfTrainImage20, SelfSupervised19, FewShot19, SelfTrainDomainAdapt19, SelfTrainSeqGen19}. Such student-teacher training step can be repeated as long as the performance improves, with the student being a teacher in the next step. Data augmentation is often used together during student-training step for further improvements \cite{SelfTrainImage20, SelfSupervised19}.

Data augmentation is another effective technique to boost model accuracy without requiring more training data. Augmenting data by adding reverberation or mixing with noise have been used in ASR (automatic speech recognition) and KWS (keyword spotting) \cite{Agc15} with some success. Recently introduced spectral augmentation \cite{SpecAug19, SpecAug20} is a new data augmentation technique shown to boost ASR accuracy significantly. In a recent work, \cite{NoisyStudentASR20} showed that applying spectral augmentation on student's input can improve ASR accuracy in self-training setup.

In this paper, we explore an application of self-training with labeled and unlabeled data where aggressive data augmentation (spec augmentation) is applied to the input of both student and teacher. The proposed student-teacher training approach enables utilization of unlabeled (unsupervised) training data for KWS problem, and also helps in applying aggressive spectral augmentation to boost diversity of training data further. 

Aggressive data augmentation can degrade accuracy in keyword spotting when used with supervised training with hard-labels ($\in\{0,1\}$). If one applies very aggressive augmentation on a positive example, one can end up with an example that may actually be seen as  negative but still has positive label. Although it may be not frequent, it can degrades the accuracy significantly by increasing false accept rate. With the proposed noisy student-student approach, teacher model generates soft-labels($\in[0,1]$) which dynamically reflects the degree of degradation in the input pattern. Supervised-training with predetermined hard-labels cannot reflect such changes of input pattern.
With experiments, we show that the proposed noisy student-teacher training with spec augmentation boosts accuracy of the model in more challenging conditions (accented, noisy and far-field). Such benefits can be explained from the use of large scale unlabeled data and aggressive data augmentation.

 We describe the proposed approach in Section \ref{sec:system}.
Then we show experimental setup in Section \ref{sec:ExperimentalSetup}, and the results in Section \ref{sec:results}. We conclude with discussions in Section \ref{sec:conclusion}.

\section{Noisy student-teacher self-training with spec augmentation}
\label{sec:system}

\subsection{Noisy student-teacher Self-training}
\label{NoisyStudentTeacherSelfTraining}

\begin{figure}
	\centering
	\includegraphics[width=\columnwidth]{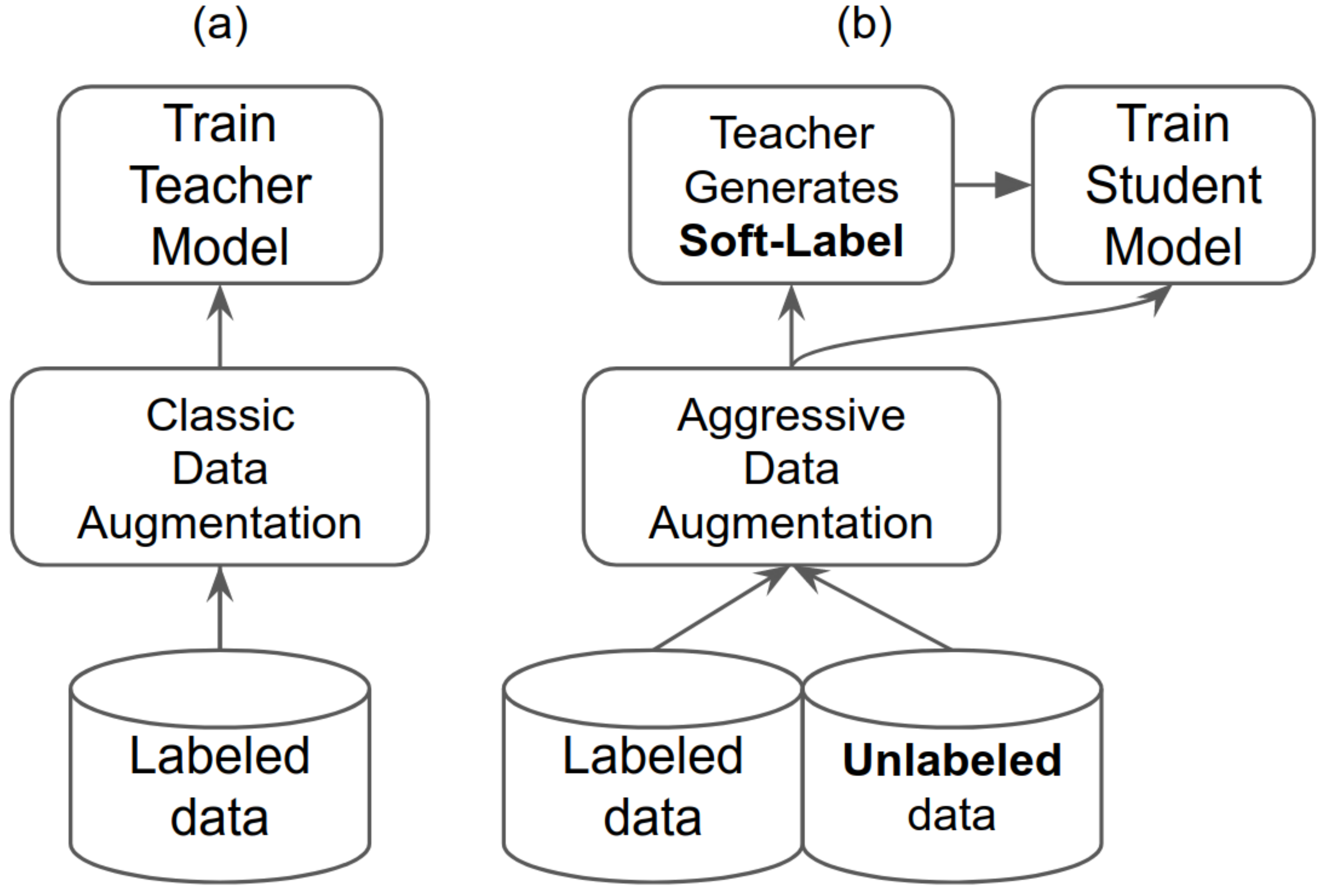}
	\caption{Self-training with Noisy student-teacher
(a) Teacher model is trained by labeled data.
(b) Student model is trained by labeled+unlabeled data with teacher logit and data augmentation.}
	\label{fig:system_concept}
\end{figure}

We propose noisy student-teacher self-training approach which consists of two major stages. In the first stage, we train a teacher model (which is also a baseline model) using conventional supervised training method on labeled data (shown in Fig. \ref{fig:system_concept} (a)). We use the same architecture and training method developed in previous work \cite{MaxPool20} for the first stage model. Also the same conventional data augmentation method (add reverberation and background noises)\cite{Agc15} is used in the first stage. The learned teacher model is passed to the second stage.

In the second stage, we train a student model using soft-labels generated from the teacher model trained in previous stage. Since the teacher provides soft-label, we can use additional unlabeled data for training student model. Also we can add more aggressive data augmentation on top of existing classical one to boost accuracy. Specifically we apply spectral augmentation which masks specific frequencies or time frames completely. Such strong modification might even change a positive pattern to a negative one, which will make an incorrect training example under supervised training method. But in student-teacher approach, the teacher can compensate for such drastic changes by generating correspondingly lower confidence. To get such benefits, both the teacher and the student model is getting the same augmented data (Fig. \ref{fig:system_concept}(b)). 

In the original self-training with noisy-student approaches \cite{SelfSupervised19, NoisyStudentASR20}, a teacher model is provided with clean data and only the student is given noisy (augmented) data. This seems to be working well for multi-class classifications such as ImageNet (objects) or ASR (graphemes) problems. But we found that providing the same noisy input to the teacher and the student achieves better performance in Keyword Spotting problem. This seems to be due to the difference of the problem, where KWS is a binary classification task with highly unbalanced pattern space. In KWS, the space of positive pattern is much smaller than that of negative patterns. Thus augmenting a positive pattern can easily result in moving the pattern into the space of negative patterns.
Also our approach is different from \cite{SelfSupervised19, NoisyStudentASR20} that both labeled and unlabeled data go through the teacher model to produce soft-labels used to train the student model. In previous works, labeled data is used for computing supervised loss on the student model while unlabeled data is used to generate soft-label. Also unlike \cite{NoisyStudentASR20}, we don't have separate data selection for the second stage.

\begin{algorithm}[ht]
	\begin{center}
		\begin{tabular}{p{83mm}}
      		 \begin{enumerate}
			\item Train Teacher $T_0$ with labeled data L and classic augmentation.
			\item Train Student $S_k$ by
      		 \begin{enumerate}
 			 \item Apply aggressive augmentation on $L \cup D$ \newline 
 			 $ D = Augment(L \cup U)$     
			 \item Use teacher $T_k$ to generate soft-label $y^T$ by \newline 
			 $ y^{T_k}(i) = f^{T_k}(x(i)) \:\:\: for \:\: x(i) \in D$    
			 \item Student model trained using CE loss by 
			 	\newline  $y^{S_k}(i)=f^{S_k}(x(i)) \:\:\: for \:\: x(i) \in D$  
			 	\vadjust{\vspace{5pt}}
				\newline  $Loss = CE(y^{T_k}, y^{S_k})$
			\end{enumerate}
			\item Set $T_{k+1} = S_k$ and Repeat step 2 
			\end{enumerate}
		\end{tabular}
		\caption{Self-training with noisy student-teacher}
		\label{alg:algorithm}
		
	\end{center}
\end{algorithm}

\begin{equation}
\label{eq:loss-eqn-combined}
 Student\_teacher\_loss = \alpha * Loss^E + Loss^D
\end{equation}

\begin{equation}
\label{eq:loss-eqn-decoder}
 Loss^D = \emph{cross\_entropy}(y^T_d, y^S_d)
\end{equation}

\begin{equation}
\label{eq:loss-eqn-encoder}
 Loss^E = \emph{cross\_entropy}(y^T_e, y^S_e)
\end{equation}

\begin{equation}
\label{eq:teacher-output}
 y^T = [y^T_d, y^T_e] = f^T(augment(x))
\end{equation}

\begin{equation}
\label{eq:student-output}
 y^S = [y^S_d, y^S_e] = f^S(augment(x))
\end{equation}

The proposed method can be summarized by Algorithm \ref{alg:algorithm}. As shown, we can also have multiple iterations (indexed by k) of the second stage by using the student from previous iteration $S_k$ as the teacher model $T_{k+1}$ for next iteration. 
Losses for student-teacher training is computed by cross entropy (Eq. \ref{eq:loss-eqn-combined}-\ref{eq:student-output}). Note that we compute two cross entropy's (for encoder and decoder labels), since our baseline model has both encoder and decoder as outputs \cite{MaxPool20, Alvarez2019}. We combine two CE losses by weighted summation (Eq. \ref{eq:loss-eqn-combined}).

\subsection{Spec Augmentation}
\label{SpecAugmentation}

Data augmentation works by generating multiple variations of an original data example using various transforms \cite{Agc15,SpecAug19,SpecAug20}, effectively multiplying number of training examples seen by the model. Classic approaches include adding reverberation or mixing with noise \cite{Agc15}. Recently proposed spectral augmentation method showed that one can boost ASR accuracy significantly by randomly masking blocks of frequency bins mostly \cite{SpecAug19,SpecAug20}. In this paper we explore the use of time and frequency masking, which is known to be most effective.

Spec augmentation is an aggressive data augmentation, in the aspect that it masks significant portion of input frequency bins or time frames in chunks. In ASR domain, such aggressive masking seems to help preventing over-fitting and facilitating the use of high level context. Also in ASR the target classes (phonemes or graphemes) are relatively well balanced in terms of prior. Meanwhile, KWS typically is a binary classification problem where positive pattern occupies only a small pattern space, while negative patterns span all the other spaces. One can easily transform a positive pattern to be a negative one by masking chunks of frequency bins. In supervised learning with predetermined hard-label, those labels can simply be incorrect after some augmentation. To overcome such over-augmentation issue, we proposes to use spec augmentation with noisy student-teacher setup.

\subsection{Model Architecture}
\label{ModelArchitecture}

For both the teacher (baseline) and the student model, we use the same two stage model architecture as in \cite{Alvarez2019, MaxPool20}. The model consists of 7 simplified convolution layers and 3 projection layers, being organized into encoder and decoder sub-modules connected sequentially. Encoder module takes the input feature which is a 40-d vector of spectral frequency energies and generates encoder output of dimension N which learns to encode phoneme-like sound units. The decoder model takes the encoder output as input and generates binary output that predicts existence of a keyword in the input stream. For more details, please refer to \cite{Alvarez2019, MaxPool20}.

\section{Experimental setup}
\label{sec:ExperimentalSetup}

\subsection{Model setup}
\label{sec:ModelSetup}

We implemented and compared the proposed model with baseline and some other variations as summarized in Table \ref{tab:models}. In the table, Baseline\_MP denotes the baseline model from our previous work \cite{MaxPool20} which uses supervised learning with labeled data (L) and max pooling loss. This model also becomes the first teacher model($T_0$) in Algorithm \ref{alg:algorithm}. Model MP+sAug is a a variant of the baseline model by simply adding spec augmentation on top of existing classic data augmentation. Model ST denotes a student-teacher trained model (using $T_0$) with classic augmentation and additional unlabeled data (U). Model ST+sAug is the same as model ST except that spectral augmentation is applied on top of classical augmentation. Model ST+sAug g2 denotes a second generation student-teacher trained model where the previous ST+sAug model was used as its teacher. ST+sAug NS(noisy student) is the same as ST+sAug model except that spec augmentation is applied only on student's input similarly to \cite{SelfTrainImage20,NoisyStudentASR20}. As shown in Table \ref{tab:models}, all student-teacher trained models are trained using both labeled and unlabeled data, while supervised training models were trained using labeled data only.

\begin{table}[th]   
	\vspace{-2mm}
	\begin{center}
		\caption{Summary of various models tested}
		\label{tab:models}
		\begin{tabular}{c|c|c}
			\textbf{Models} & Loss & Training data\\
			\hline
			Baseline\_MP & MaxPool CE Loss \cite{MaxPool20}  & L \\
			MP+sAug    & MaxPool + sAug  &   L  \\
			\hline
			ST         &  ST (Student-Teacher) & L + U \\
			ST+sAug    &  ST + sAug   & L + U \\
			ST+sAug g2 &  ST + sAug 2nd gen. & L + U \\
			ST+sAug NS &  ST with noisy student + sAug & L + U \\
		\end{tabular}
	\end{center}
	\vspace{-8mm}
\end{table}

\subsection{Training data set}
\label{dataset}

We used both supervised (labeled) training data, and unsupervised (unlabeled) training data for experiments.
Our supervised training data consists of 2.5 million anonymized utterances with the keywords (“Ok Google” or “Hey Google”). Supervised data is labeled by large ASR model similarly to \cite{Alvarez2019, MaxPool20}. The unsupervised training data consists of 10 million anonymized utterances with the keywords and noises. The unsupervised data has relatively high noise level making it difficult for ASR model to generate reliable labels.

\subsection{Evaluation data set}
\label{ssec:data}

Evaluation is done with 6 positive data sets separate from training data, where each set represents a diverse environmental condition as summarized in Table \ref{tab:EvalSet}. In the table, QLog, QLog diff, QLog easy are anonymous query logs from different time period and conditions. Especially QLog-low has utterances that score relatively low confidences (difficult to detect), while Qlog-high has utterances that score relatively high confidences (easy to detect). 
\begin{table}[h!]
	\vspace{-4mm}
	\begin{center}
		\caption{Summary of evaluation dataset}
		\label{tab:EvalSet}
		\begin{tabular}{c|l}
			\textbf{Eval set name} & Description (\# of utt's) \\
			\hline
			Near Cl      & Nearfield Clean (170K)   \\ 
			Near Cl Acc  & Nearfield Clean Accented(16k)  \\
			Far Cl       & Farfield Clean  (1k)    \\
			Far Mus      & Farfield w/ music noise (1k) \\
			Far TV       & Farfield w/ TV noise (1k)  \\
			QLog         & Anonymous query logs (87k)  \\
			QLog low     & Qlog's with lower confidence (265k)  \\
			QLog high    & Qlog's with higher confidence (255k) \\
		\end{tabular}
	\end{center}
	\vspace{-6mm}
\end{table}

\section{Results}
\label{sec:results}

\begin{table}[h!]
	\vspace{-4mm}
	\begin{center}
		\caption{FR rate of models with various loss types at 0.1 FA/h}
		\label{tab:perfdiffA}
		\begin{tabular}{l|r|r|r|r}
			\textbf{Models} & Near Cl  & Far Cl & Far Mus & Far TV    \\
			\hline
			Baseline\_MP   & 0.56\% & 1.83\% & 15.18\%  & 27.94\% \\
			MP+sAug        & 1.17\%  & 6.28\% & 32.24\% & 47.96\% \\
			\hline
			ST             & 0.53\%  & 1.57\% & 15.06\% & 27.58\%  \\
			ST+sAug NS     & 0.74\%  & 1.05\% & 17.65\% & 30.22\%  \\
			ST+sAug        & 0.59\%  & 0.96\% & 12.00\% & 24.58\%  \\
			ST+sAug g2     & 0.53\%   & 0.78\% & 13.65\% & 25.06\%  \\
			\hline  			  			
 			\hline
			\textbf{Models} & NearClAcc & QLog & QLog low & QLog high   \\
			\hline
			Baseline\_MP   & 1.64\%  & 8.21\% & 11.07\% & 1.73\%  \\
			MP+sAug        & 3.38\% & 18.74\% & 27.16\% & 8.48\%  \\
			\hline
			ST             & 1.58\% & 6.00\% & 8.63\% & 1.24\% \\
			ST+sAug NS     & 2.38\% & 7.38\% & 11.57\% & 2.14\% \\
			ST+sAug        & 1.71\% & 3.83\% & 7.28\% & 0.99\% \\
			ST+sAug g2     & 1.42\% & 3.12\% & 7.15\% & 0.91\%  \\
		\end{tabular}
	\end{center}
	\vspace{-4mm}
\end{table}

\begin{figure}[htb]
	\begin{minipage}[b]{\linewidth}
		\centering
		\centerline{\includegraphics[height=137pt]{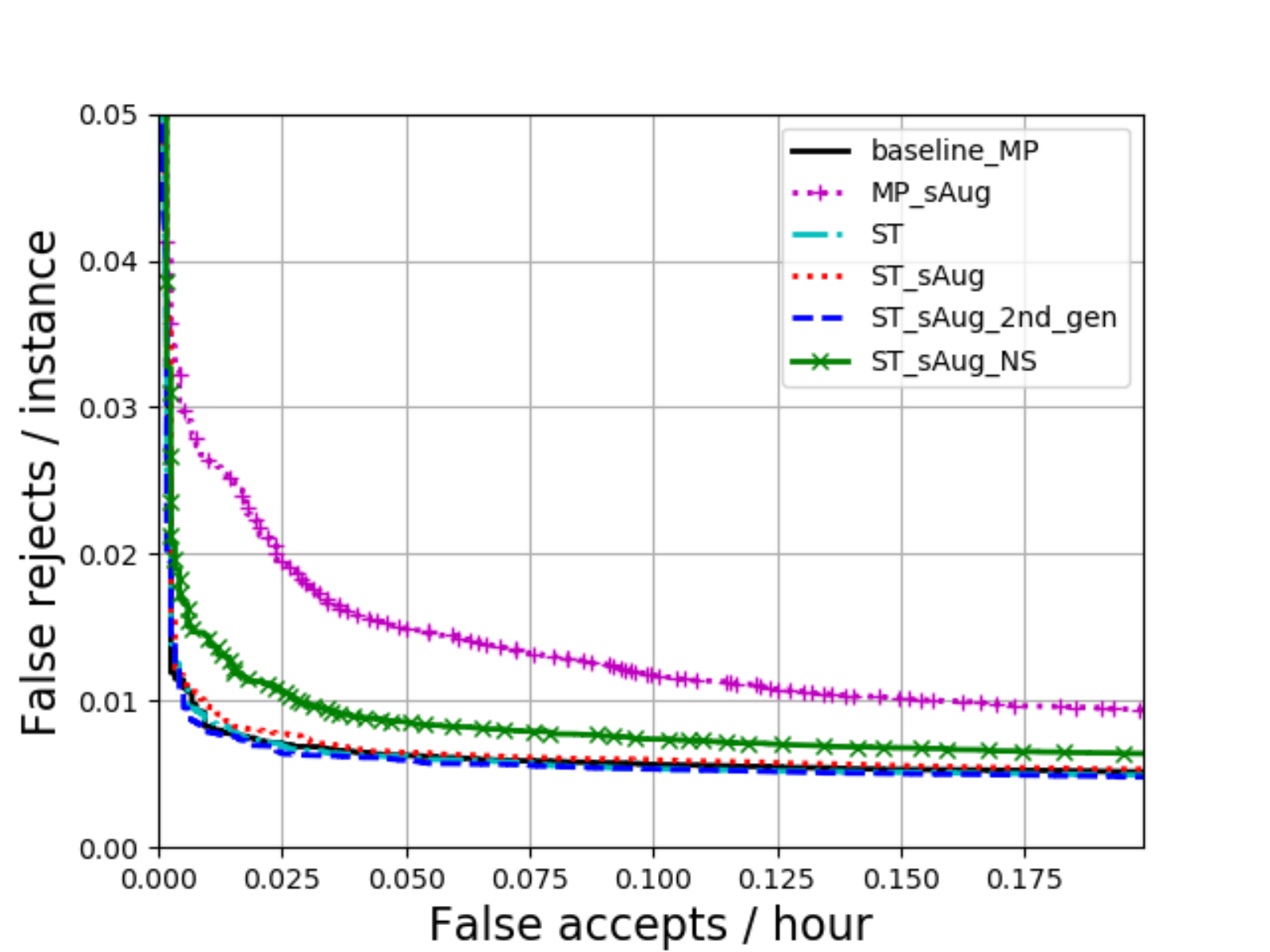}}
		\centerline{(a) Near-field Clean }\medskip
	\end{minipage}
	\hfill
	\begin{minipage}[b]{\linewidth}
		\centering
		\centerline{\includegraphics[height=137pt]{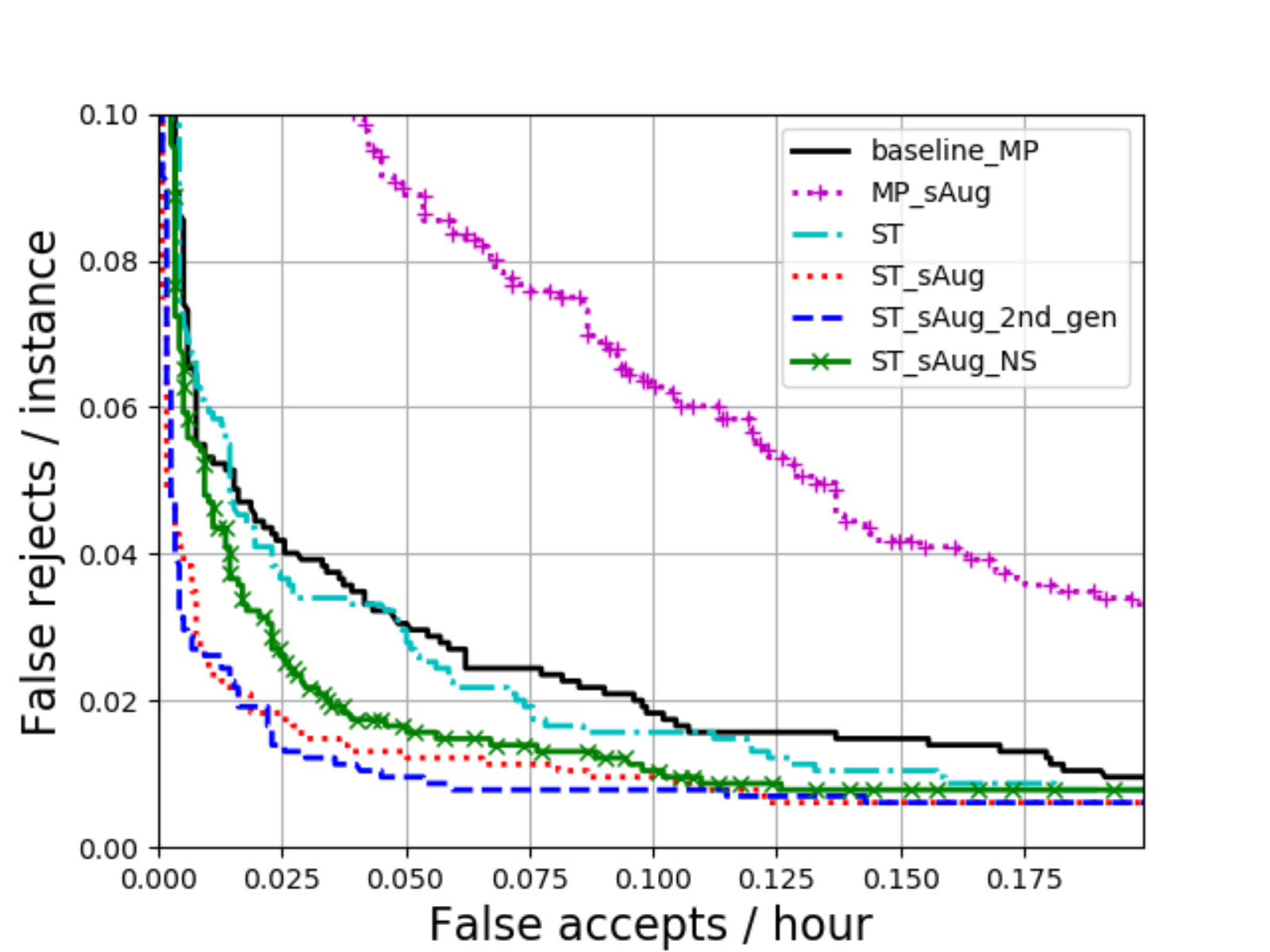}}
		\centerline{(b) Far-field Clean}\medskip
	\end{minipage}
	\begin{minipage}[b]{\linewidth}
		\centering
		\centerline{\includegraphics[height=137pt]{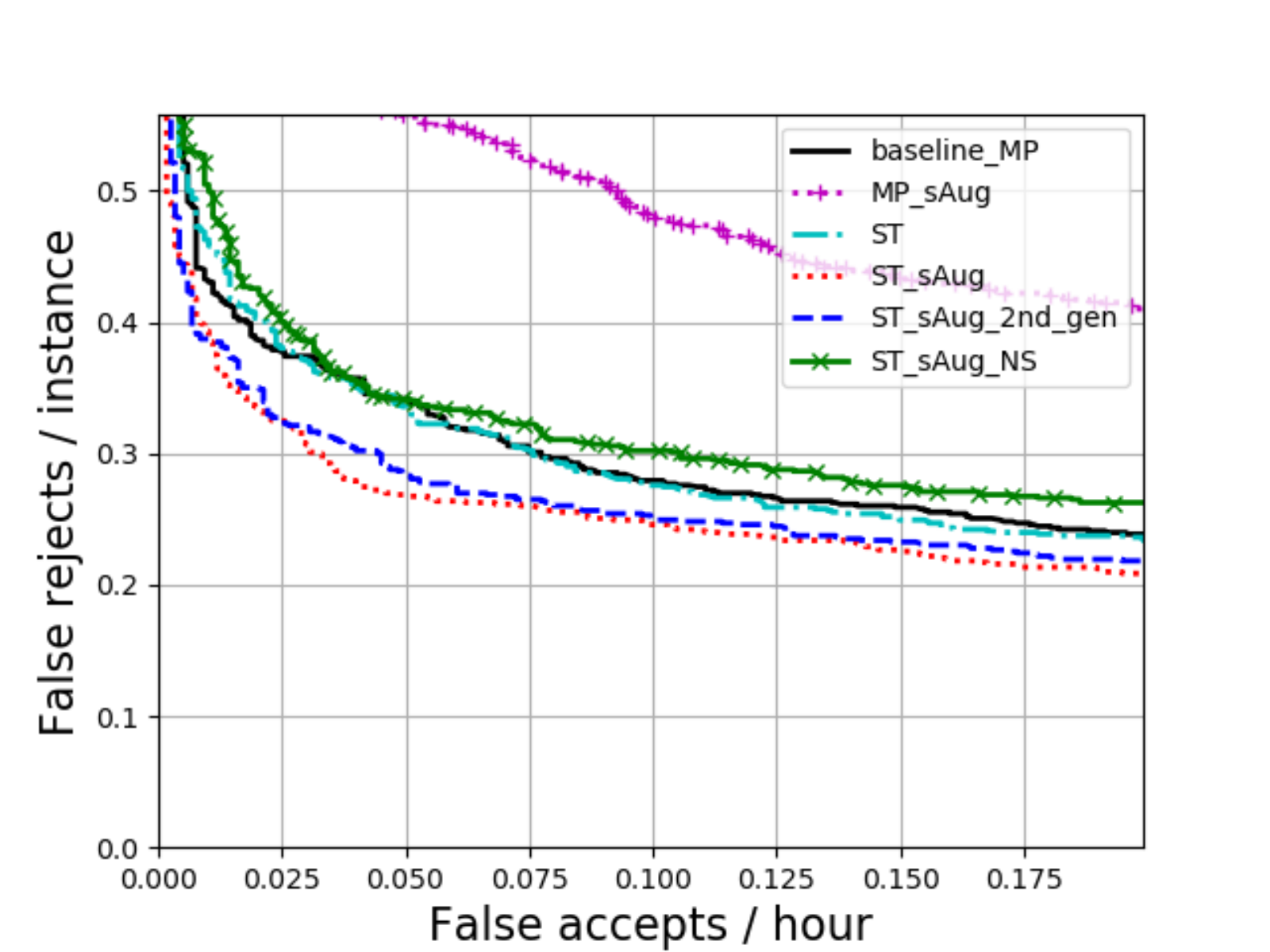}}
		\centerline{(c) Far-field with TV noise}\medskip
	\end{minipage}
	\hfill
	\begin{minipage}[b]{\linewidth}
		\centering
		\centerline{\includegraphics[height=137pt]{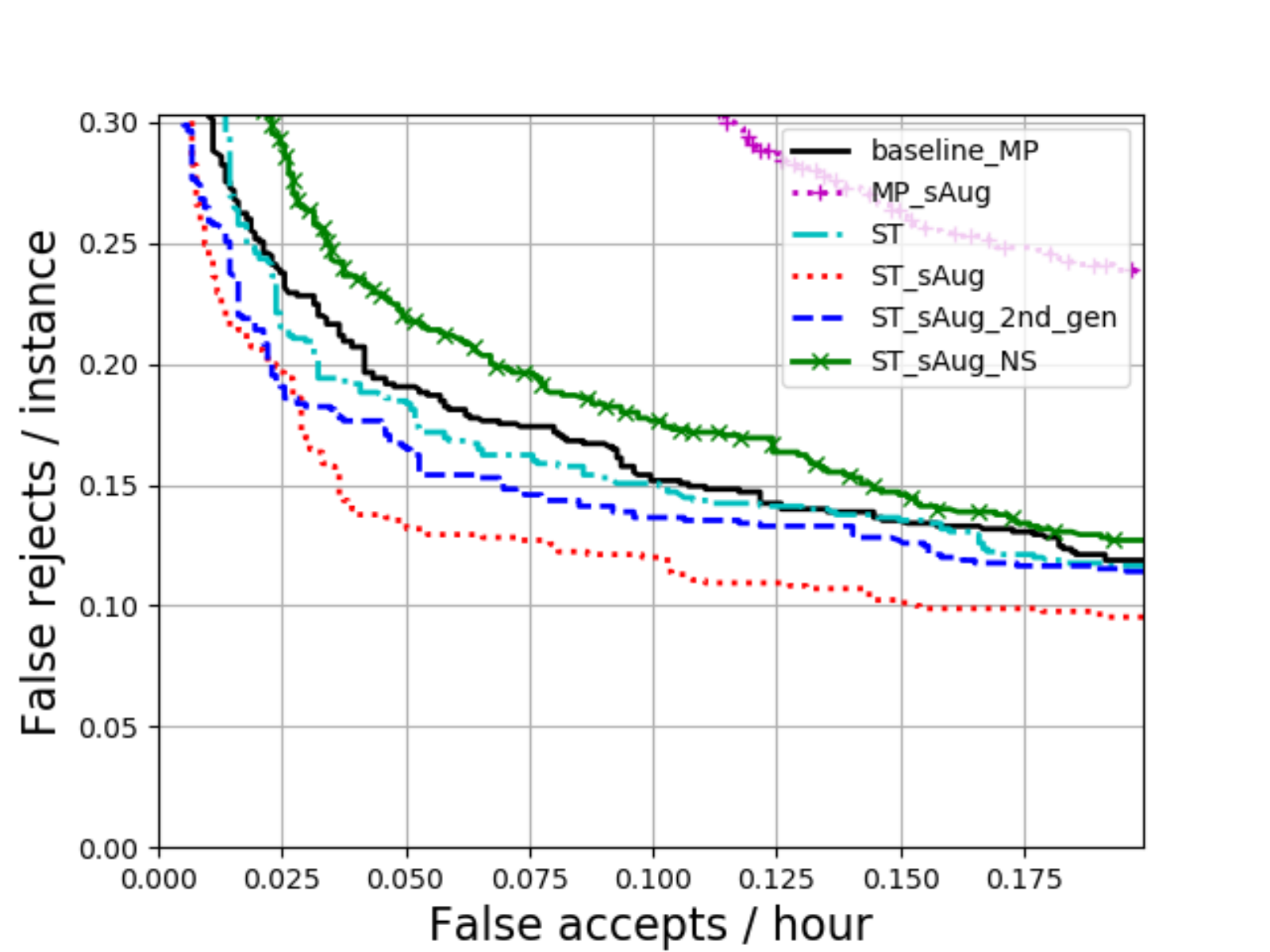}}
		\centerline{(d) Far-field with Music noise}\medskip
	\end{minipage}
	\caption{ROC curves of models with various recipes and conditions}
	\label{fig:rocA}
\end{figure}

\begin{figure}[htb]
	\begin{minipage}[b]{\linewidth}
		\centering
		\centerline{\includegraphics[height=137pt]{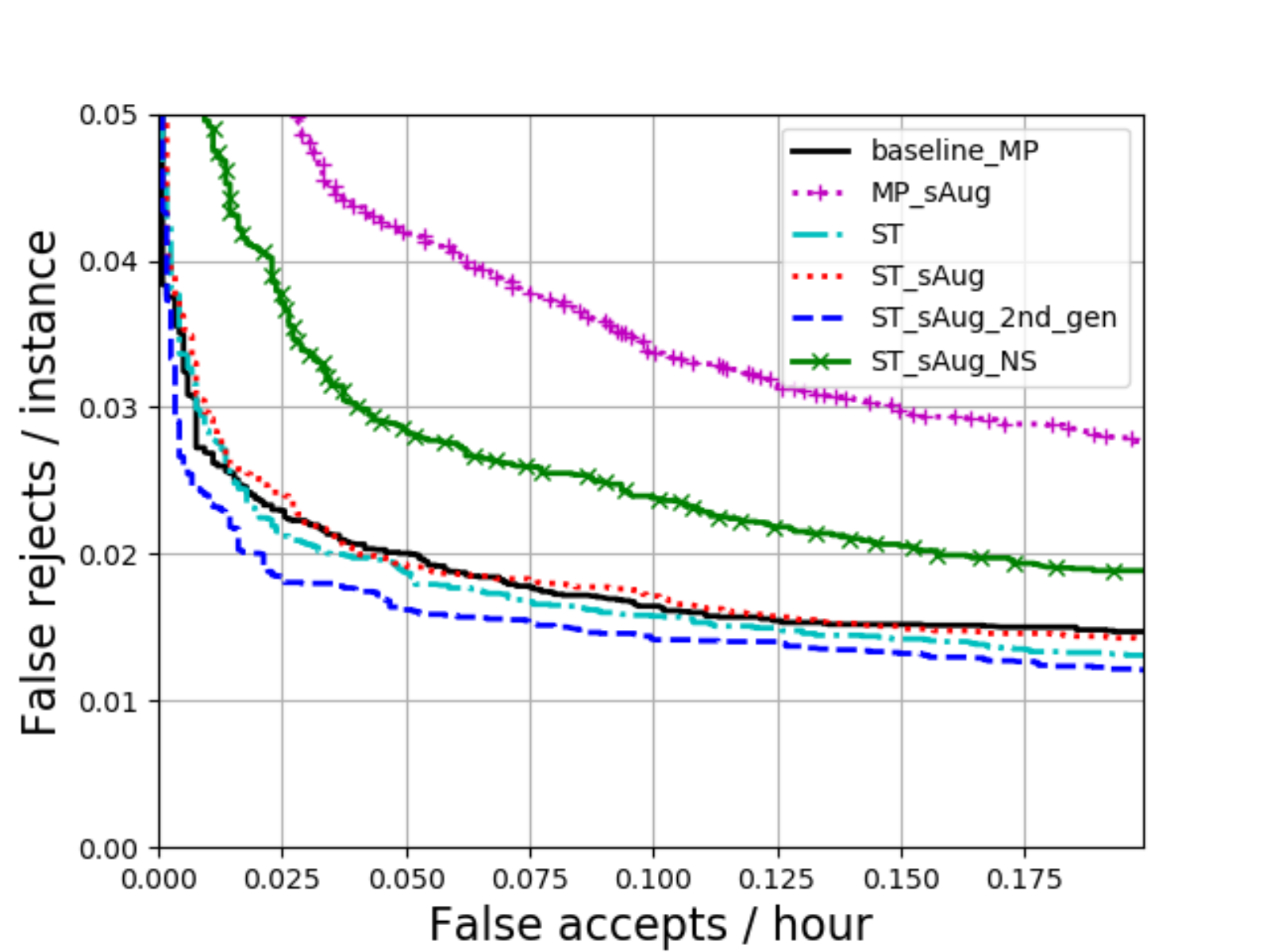}}
		\centerline{(e) Near-field Clean Accented }\medskip
	\end{minipage}
	\hfill
	\begin{minipage}[b]{\linewidth}
		\centering
		\centerline{\includegraphics[height=137pt]{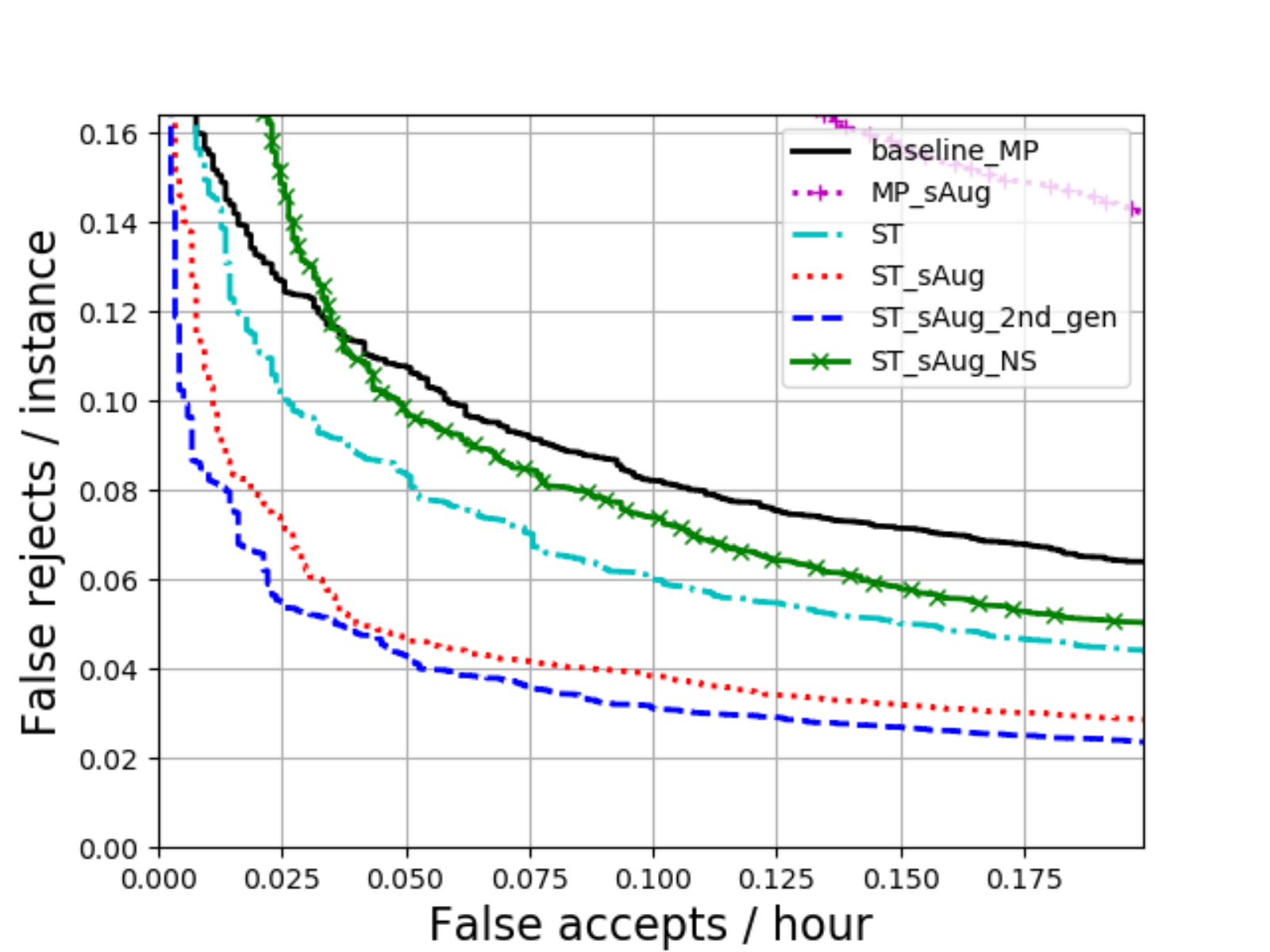}}
		\centerline{(f) Anonymous query Logs}\medskip
	\end{minipage}
	\hfill
	\begin{minipage}[b]{\linewidth}
		\centering
		\centerline{\includegraphics[height=137pt]{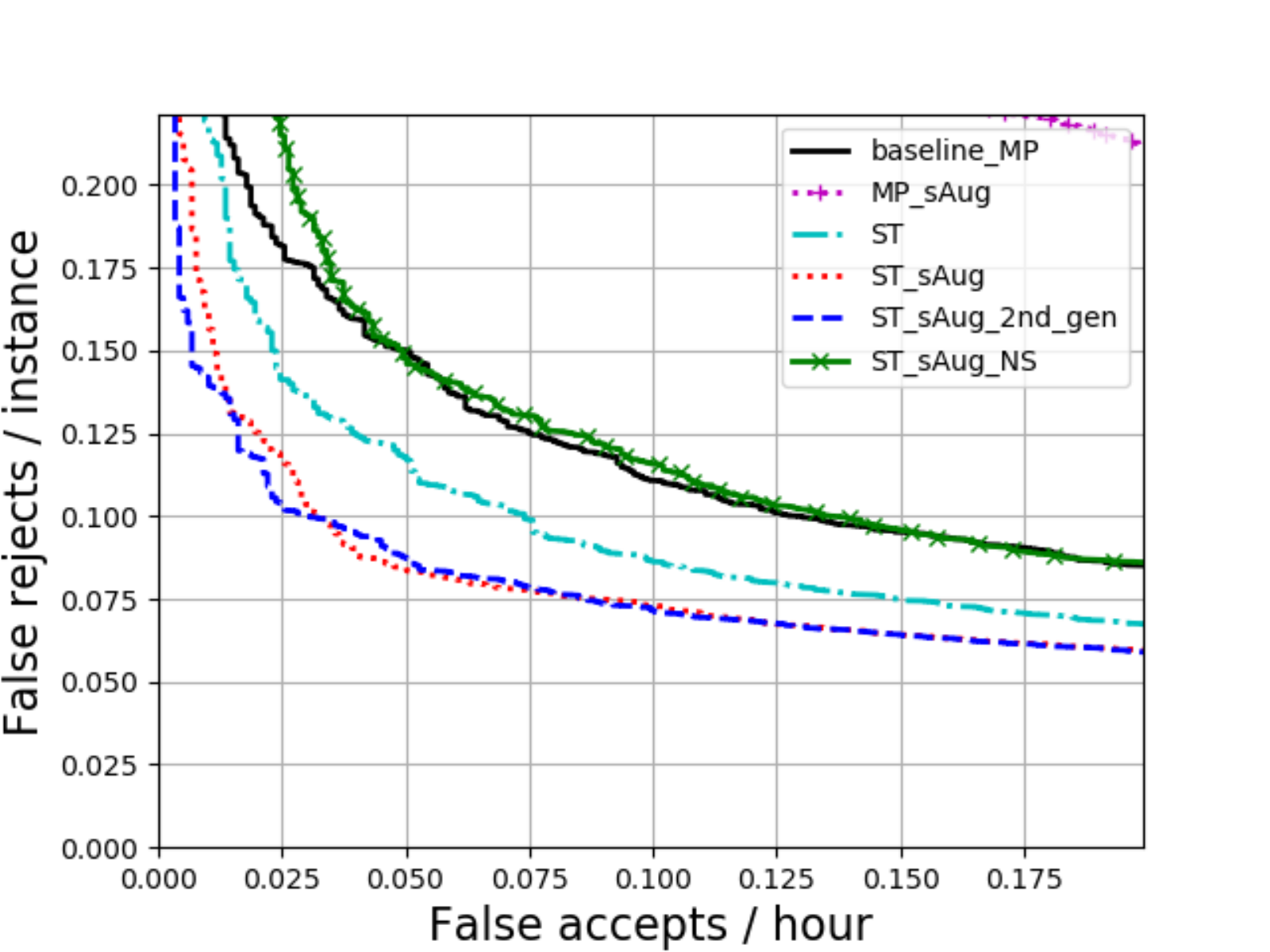}}
		\centerline{(g) Anonymous query Logs with low confidence}\medskip
	\end{minipage}
	\begin{minipage}[b]{\linewidth}
		\centering
		\centerline{\includegraphics[height=137pt]{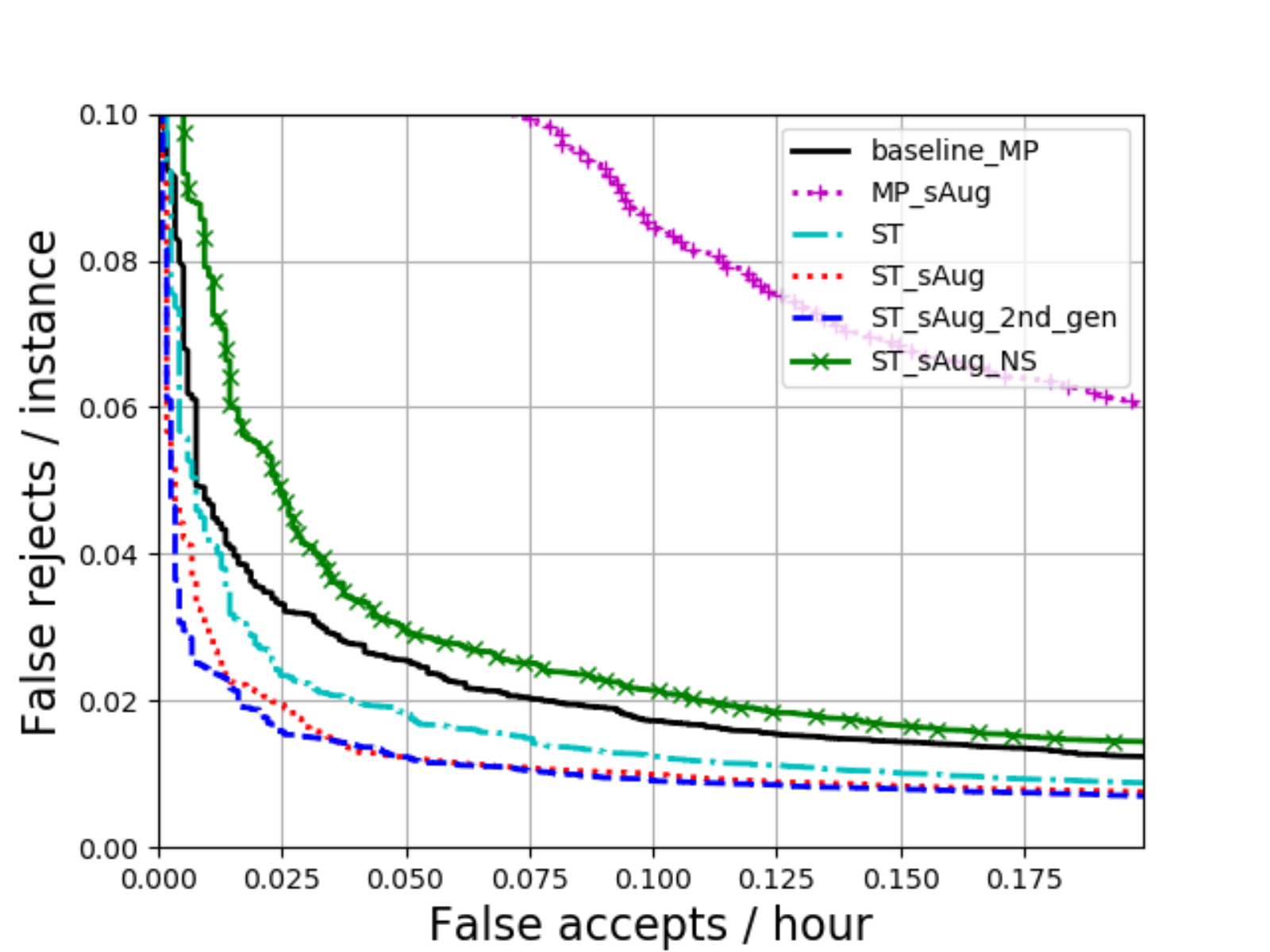}}
		\centerline{(h) Anonymous query logs with high confidence}\medskip
	\end{minipage}
	%
	\caption{ROC curves of models with various recipes and conditions}
	\label{fig:rocB}
\end{figure}

We evaluated 6 types of trained models on 8 evaluation sets and results are summarized by Tables and figures. Table \ref{tab:perfdiffA} summarizes FR rates of the models at selected FA/h rate (0.1 FA per hour measured on 64K re-recorded TV noise set). Fig.\ref{fig:rocA} and \ref{fig:rocB} shows the ROC (receiver operator characteristic) curves of tested models across different evaluation sets. 

Results show that simply applying spec augmentation on top of classic augmentation with supervised training (MP+sAug model) doesn't work well with our setup. This seems to be due to the risks of over augmentation (that transforms a positive labeled example into negative example with positive label).

The proposed model (ST+sAug and ST+sAug g2) showed significant improvements over baseline for difficult conditions such as Far-field and Query Logs. For example, Far-field Clean condition improved from 1.83\% (baseline) to 0.78\% (ST+sAug g2). Query Logs condition improved from 8.21\% (baseline) to 3.12\% (ST+sAug g2) (60\% relative improvement). Clean Accented condition also improved from 1.64\% to 1.42\% (ST+sAug g2). ROC plots (Fig \ref{fig:rocA}, \ref{fig:rocB}) shows similar trends. Simple student-teacher training (ST model) shows small improvements over baseline (teacher model), assisted by extra unlabeled data. But the improvements are relatively minor compared to ST+sAug or ST+sAug g2.

Direct adaptation of noisy student self-training method (Model ST+sAug NS) didn't work well on our KWS problem setting. In this model, we apply spec augmentation on only the student model's input, and not on the teacher model. Similarly to the MP+sAug model, the aggressive augmentation can transform positive example to a negative one, while the teacher model doesn't see such changes resulting in incorrect soft-labels.

\section{Conclusion}
\label{sec:conclusion}
We presented self-training with noisy student-teacher for keyword spotting problem. The proposed approach enables the use of abundant unlabeled data and aggressive augmentation. Experimental results show that models with proposed approach significantly improves on evaluation set with difficult conditions. Experiments also show that applying aggressive augmentation directly in supervised learning approach doesn't work well for keyword spotting problem, while semi-supervised training with noisy student-teacher can benefit from aggressive augmentation and unlabeled data.

\clearpage
\bibliographystyle{IEEEtran}
\bibliography{refs}

\end{document}